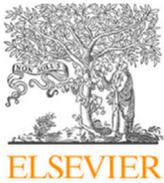
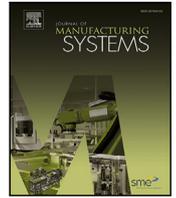

Technical paper

# A sensor-to-pattern calibration framework for multi-modal industrial collaborative cells☆

Daniela Rato [a,b,*], Miguel Oliveira [a,b], Vítor Santos [a,b], Manuel Gomes [b], Angel Sappa [c,d]

[a] *Institute of Electronics and Informatics Engineering of Aveiro, University of Aveiro, Campus Universitário de Santiago, 3810-193 Aveiro, Portugal*
[b] *Department of Mechanical Engineering, University of Aveiro, Campus Universitário de Santiago, 3810-193 Aveiro, Portugal*
[c] *Computer Vision Center, Edificio O, Campus UAB, Barcelona, 08193 Bellaterra, Spain*
[d] *Escuela Superior Politécnica del Litoral, Campus Gustavo Galindo - 2019, Guayaquil, Ecuador*

A R T I C L E   I N F O

A B S T R A C T

Collaborative robotic industrial cells are workspaces where robots collaborate with human operators. In this context, safety is paramount, and for that a complete perception of the space where the collaborative robot is inserted is necessary. To ensure this, collaborative cells are equipped with a large set of sensors of multiple modalities, covering the entire work volume. However, the fusion of information from all these sensors requires an accurate extrinsic calibration. The calibration of such complex systems is challenging, due to the number of sensors and modalities, and also due to the small overlapping fields of view between the sensors, which are positioned to capture different viewpoints of the cell. This paper proposes a sensor to pattern methodology that can calibrate a complex system such as a collaborative cell in a single optimization procedure. Our methodology can tackle RGB and Depth cameras, as well as LiDARs. Results show that our methodology is able to accurately calibrate a collaborative cell containing three RGB cameras, a depth camera and three 3D LiDARs.

## 1. Introduction

According to the European Commission, Industry 5.0 aims to strengthen the contribution of the industry to society, by thinking beyond efficiency and productivity, aiming the development of technology towards the improvement of the worker's quality of life, while also respecting the planet [1,2]. With this expansion of Industry 5.0, many new technologies are arriving to facilitate industrial and manufacturing jobs for humans by removing heavy burdens such as lifting of heavy weights and repetitive movements. For that matter, collaboration with robots has become a highly researched topic because it can combine the expertise of humans with the workload of a robot [3,4].

A collaborative cell is a tridimensional space where a collaborative robot and humans can safely coexist and perform common tasks. Safety requirements in collaborative setups are standardized [5] with strict restrictions when it comes to robot motion during human collaboration. These restrictions are mainly related to speed or torque limitations when the robot is operating close to humans. These rigorous requirements demand a robust perception framework inside the cell, especially around the robot. This can only be achieved with a wealth of sensors positioned in strategic places to cover the cell and overcome potential occlusions that are bound to happen with the movements of people, objects, and the robot itself. In addition to the multi-sensor setup, having also a multi-modal sensor system brings complementary information that can ensure a higher level of safety. For example, range data (from LiDARs and RGB-D cameras) can be used for volumetric monitoring, and RGB data (from cameras) can be used for object detection. Furthermore, the data can even be fused for human pose estimation, which can ensure that the position of the humans inside the cell is known at all times, guaranteeing their security.

Unfortunately, in spite of the great advantage, data fusion is a complex problem, and the challenge is to fuse the data of several multi-modal sensors to create a single intelligent system in the collaborative cell. According to [6], the challenges of multi-modality can be divided into five major categories: representation, translation, alignment, fusion, and co-learning. Representation refers to how multi-modal data is

☆ This work was supported in by the Foundation for Science and Technology (FCT) under the grant 2021.04792.BD. The present study was also developed in the scope of the Project Augmented Humanity [POCI-01-0247-FEDER-046103], financed by Portugal 2020, under the Competitiveness and Internationalization Operational Program, the Lisbon Regional Operational Program, and by the European Regional Development Fund. The authors acknowledge the support of CYTED Network: Ibero-American Thematic Network on ICT Applications for Smart Cities (REF-518RT0559).
* Corresponding author at: Department of Mechanical Engineering, University of Aveiro, Campus Universitário de Santiago, 3810-193 Aveiro, Portugal.
*E-mail addresses:* danielarato@ua.pt (D. Rato), mriem@ua.pt (M. Oliveira), vitor@ua.pt (V. Santos), manuelgomes@ua.pt (M. Gomes), asappa@cvc.uab.cat (A. Sappa).

https://doi.org/10.1016/j.jmsy.2022.07.006
Received 8 June 2022; Received in revised form 1 July 2022; Accepted 17 July 2022
Available online 9 August 2022

structured, exploring the data's complementary and redundancy. Translation refers to mapping data from one modality to another. Alignment is the identification of direct relations between elements/sub-elements from different modalities. Fusion is the grouping of several modalities to perform predictions. Furthermore, co-learning treats the transference of knowledge between modalities.

So, one can conclude that determining the alignment between sensors is an essential task when trying to fuse multi-modal information, leaving us with the problem of extrinsic calibration: the process of determining the transformation between a set of sensors. There are two main concerns associated with the calibration of the sensors in a collaborative cell. First of all, the safety demands mentioned above also require that this calibration is accurate, which is not trivial with such complex systems. Secondly, the number of sensors and their different modalities create amounts of data too large to be processed simultaneously. The multi-modal characteristics of these systems also raise the challenge of processing and combining data from entirely different modalities. An example of this is selecting a calibration target that all modalities are able to detect.

There are already solutions that tackle the extrinsic calibration problem. However, those solutions cannot tackle the collaborative cell problem that brings additional calibration challenges. Examples of these are a multi-sensor and multi-modal setup, the requirement of high accuracy solutions, and the high density of data produced. Also, the cell can be significantly large, meaning that the field of view of the placed sensors may not entirely overlap, already eliminating the most common techniques like the *OpenCV* calibration tool, which requires cameras to have overlapping fields of view.

Our approach tackles these challenges by proposing a calibration framework based on optimizing sensor-to-pattern transformations, which can calibrate complex robotic systems and handle RGB, depth and LiDAR modalities. Unlike most methodologies that use a sensor-to-sensor approach, we propose a method that uses a sensor-to-pattern approach, which makes the calibration largely simplified. Because of this, our framework can also handle sensors with non-coincident fields of view.

The contributions of this paper are:

- The development of a calibration framework able to calibrate complex, multi-modal and multi-sensor setups
- A solution to calibrate sensors with non-overlapping fields of view
- A calibration framework able to calibrate RGB, LiDAR and depth modalities

The remainder of this paper describes and proves the concept of the methodology undertaken to tackle the extrinsic calibration problem in a generic way. In Section 2, the authors explore the different calibration approaches in the literature and compare them to the one developed. Section 3 describes the process of data acquisition, labeling and calibration itself. Section 4 presents the results obtained in our collaborative cell with a robotic system of 7 sensors with three different modalities and a comparative study with other approaches in the literature. Finally, Section 5 summarizes the contributions of this paper and exposes what could be done to improve the methodology in future works.

## 2. State of the art

As discussed in Section 1, to adequately monitor a complex surrounding environment, such as a collaborative cell, robotic systems need to process multi-modal information from multiple sensors. In addition to this, it is also clear that processing all this information streaming from various sources requires that the alignment between the sensors is previously determined. Whenever an intelligent or robotic system comprises two or more sensors, a procedure that estimates the geometric transformations between those sensors is required. The process by which these transformations are estimated is called extrinsic calibration.

The vast majority of sensor fusion techniques operate under the assumption that accurate geometrical transformations between the sensors that collect the data are known. This is valid for many different applications, from simple cases, such as the simple design of sensors that collect RGB and depth information [7] or a stereo camera pair designed to carry out underwater 3D reconstruction [8], to more complex sensor setups such as intelligent vehicles [9,10], smart camera networks [11], robot based sensing approaches [12], or even multi-sensor image analysis from datasets captured by diverse airborne or spaceborne sensors [13]. Thus, it is clear that an accurate estimation of those transformations, i.e., a good extrinsic calibration, is a critical component of any data fusion methodology.

The majority of works in the literature focuses on the pair-wise calibration of sensors, usually RGB-RGB [14–18], RGB-depth [19–22] or RGB-LiDAR [23–30]. However, these frameworks are not adequate to calibrate collaborative cells because their extension to multi-sensor configurations is not trivial. For example, RGB-RGB pairwise calibration methodologies are not easily adapted to systems containing several cameras.

When calibrating a system with a large number of sensors, the pair-wise calibration methods require that the transformations are calculated between each combination of pairs in the system. One of the problems of pair-wise calibration is that it escalates with the number of sensors. This type of calibration also raises the question of the selection of the pairs. If we want to calibrate sensors A, B and C, and we want to get the transformation from A to C, we have multiple paths that can be chosen, for example, A-C, A-B-C or B-A-C. Moreover, we have no guarantee of which one has the lower error, especially if A and C are from different modalities and the objective function is not symmetric. There is also the problem of the order of the pair being calibrated. For example, if we have sensors A and B, the pair could be calibrated from A to B or B to A. The pair-wise calibration methods also increase the error by accumulating the error of the different combinations of pairs.

In terms of a multi-modal calibration, most techniques only use two modalities. For example, Rodriguez et al. [31] calibrate RGB cameras and multi-layer LiDARs to estimate the extrinsic parameters using on a minimum of six poses. Furgale et al. in [32] use optimization techniques for jointly estimating the temporal offset between measurements of RGB cameras and IMUs and their spatial displacements concerning each. Xu et al. in [33] estimate the relative pose for each pair of depth and thermal frames by minimizing the objective function that measures the temperature consistency between a 2D infrared image and the reference 3D thermographic model. Yang et al. in [34] propose a method to calibrate an opti-acoustic imaging system combining a camera and an FLS in underwater environments. There also are methods that use more than two modalities, as in [35], where Pereira et al. use classical segmentation and fitting techniques to generate sets of centers of a ball in motion in front of a set of RGB, LiDAR and depth sensors. In [36], Taylor et al. proposed a method to calibrate RGB cameras, LiDARs and IMUs that utilize the system's motion to estimate the pose of each sensor. Domhof et al. [37] use optimization to calibrate radar, RGB cameras and LiDARs. It is unusual to see techniques applied to a large variety of modalities, which is also the case of a collaborative cell that requires different levels of perception that can only be achieved with a multi-modal setup.

Additionally, there are not many methodologies that can calibrate systems where the sensors have non overlapping fields-of-view or partially overlapping fields-of-view where the overlapping is not close enough to detect a calibration pattern that is in the same field of view. An example of a methodology that solves this problem can be found in [38]. Nonetheless, the framework in [38] uses a pair-wise approach to calibrate RGB, depth cameras and range finders and requires that the sensor setup is moved in space, which, once again, does not apply to a collaborative cell, where the sensors must be fixed to the structure.

To conclude, to the best of our knowledge there is no other calibration framework that is able to calibrate such a complex system in terms of multi-modality and number of sensors with accurate results. Also, due to the size of the collaborative cell, it is very difficult to have the calibration pattern detected by all sensors at the same time without moving the sensors, which is also not solved in the state of the art methodologies.

In this paper, we focus on the problem of calibrating a highly complex system such as a collaborative cell. As discussed before, such systems require a vast amount of sensors to avoid occlusion. To ensure safety, these sensors are positioned to view different regions of the workspace, which in turn creates non-globally shared overlapping fields of view among the sensors in the system. In addition to this, collaborative cells also require multi-modal systems. To address this challenge, we integrate three commonly used sensor modalities in the calibration framework, RGB, LiDAR and depth. We have described our calibration system in other works, but focusing on distinct points: in [39] we focused on the calibration of an autonomous vehicle, while [40] describes the integration of the LiDAR sensor in the calibration framework, and finally, [41] shows how the framework may be used to calibrate hand-eye robotic systems.

## 3. Methodology: Automatic calibration

The standard procedure for calibrating multi-sensor systems is to use a calibration pattern, which is positioned in such a way that it is accurately detected by all sensors. Then, the classic approach is to formulate the calibration as an optimization procedure that minimizes a set of errors. These errors are computed by an objective function, which is designed to translate the quality of alignment between the sensors, given the transformation between those sensors.

The issue is that the errors are computed as a function of pairs of sensors. One example is the usage of the reprojection error ($e$) for calibrating multiple camera systems, expressed in Eq. (1):

$$\underset{\{{}^{s_i}\hat{\mathbf{T}}^{s_j}\}}{\arg\min} \sum_{\mathbb{S}} \sum_{\mathcal{I}} e\left({}^{s_i}\hat{\mathbf{T}}^{s_j}, d_{s_i}, d_{s_j}, \lambda_{s_i}, \lambda_{s_j}\right), \quad (1)$$

where ${}^{s_i}\hat{\mathbf{T}}^{s_j}$ is the estimated transformation between sensors $s_i$ and $s_j$, $\mathbb{S}$ represents the set of pair-wise combinations of the sensors in the system, $\mathcal{I}$ represent the set of images used to calibrate the system, $d$ denotes the detections of the pattern by a sensor, and finally $\lambda$ represents the intrinsic parameters of the sensor,

However, in the case at hand, a very complex system, the usage of errors derived from pair-wise combinations of sensors (represented by $\mathbb{S}$ in (1)) is not scalable, since one must develop different mechanisms i.e., different versions of the objective function $e$, for each pair-wise combination of modalities. Moreover, in a collaborative cell, there will be many pairs of sensors that do not overlap. In these cases, it would not be possible to compute the errors using this problem formulation.

The differentiating aspect of our calibration methodology w.r.t. to others is that it uses a sensor-to-pattern approach instead of the classic sensor-to-sensor error estimation. Since that each sensor views the pattern as a function of its intrinsic properties, pose, and pose of the pattern, instead of using the pair-wise transformation error to optimize the transformations between sensors, we estimate the error by defining a function that uses the transformation between each sensor and the calibration pattern, as expressed in Eq. (2).

$$\underset{\{{}^{s_i}\hat{\mathbf{T}}^{w}\},\{{}^{w}\hat{\mathbf{T}}^{p}_c\}}{\arg\min} \sum_{S} \sum_{C} e\left({}^{s_i}\hat{\mathbf{T}}^{w}, {}^{w}\hat{\mathbf{T}}^{p}_c, d_{s_i}, \lambda_{s_i}\right), \quad (2)$$

where ${}^{s_i}\hat{\mathbf{T}}^{w}$ is the estimated transformation between sensor $s_i$ and the world coordinate frame $w$, ${}^{w}\hat{\mathbf{T}}^{p}_c$ is estimated transformation between the pattern $p$ and $w$, which varies according to each collection $c$. Since our calibration framework tackles multi-modal systems, we refer to a moment in time where data from all sensors in the system is collected as a *collection* $c$ and not an image $i$, to account for the fact that it may contain data from other modalities. To calibrate we use a set of collections $C$ to which we refer to as a dataset. In this case, the overall error is computed by summing up the contributions of the sensors in the set of sensors $S$, as opposed to the set of pair-wise combinations of the sensors $\mathbb{S}$.

One advantage of our methodology is that only one mechanism per modality must be designed, as opposed to one mechanism per pair-wise combination of modalities. Another advantage is that it is possible to estimate an error for each sensor, provided that it views the calibration pattern. This approach is much better suited to tackle calibration systems with several non-overlapping fields of view. Because we now use the transformation between the sensors and the pattern to estimate the errors, the pose of the pattern must also be included as a parameter to be optimized ${}^{w}\hat{\mathbf{T}}^{p}_c$. As such, our calibration system estimates not only the pose of the sensors but also the pose of the calibration pattern. This may sound counter-intuitive, but in fact, by enlarging the optimization problem, we reduce its complexity.

In order to ensure robustness, the optimization should consider errors from multiple viewpoints, i.e., the sensors should observe the pattern from different viewpoints. This is a standard requirement of any calibration procedure. For example, for calibrating a stereo system, several images of both cameras are used.

In each collection we store not only the raw sensor data but also the labels that describe where the pattern was detected in the sensor data. Naturally, the detection of the pattern may fail in some cases. This may be caused by issues in the detection algorithm, such as sensitivity to illumination, or simply because the sensor does not view the pattern in that collection. When, in a collection, there is at least one sensor that does not detect the pattern, we refer to it as an incomplete collection. It is also possible that only a portion of the pattern is identified, which occurs primarily due to a partial view of the calibration pattern in RGB images. We refer to these cases as partial detections. Since a collaborative cell is a large tridimensional space and the goal is to monitor its complete volume, it is common to have small or non-existent overlapping fields of view between different sensors. It is often very difficult to find a position of the calibration pattern which is viewed by all sensors simultaneously. For that reason, the number of incomplete collections and partial detections is larger than usual in a collaborative cell system. A sensor to pattern paradigm is clearly much more adequate to tackle such complex multi-modal and multi-sensor systems.

The next sections describe the configuration of a calibration procedure and the automatic labeling and manual annotation mechanisms which are available. Finally, we detail the objective functions for each of the three presented modalities.

### 3.1. Setup and data acquisition

Since the goal is to calibrate complex robotic systems, a prior step is required for configuring the calibration. This step defines which sensors are to be calibrated. The coordinate frames in the system are hierarchically organized in a topological tree-like structure called transformation tree. The transformation tree of the collaborative cell used in this work is shown in Fig. 1.

The calibration of a sensor requires the definition of which specific transformations that is to be changed during the optimization, in order to assess if the error is minimized. This transformation must belong to the chain of transformations that go from the common reference frame (world) to the sensor's coordinate frame. For example, in Fig. 1, sensor $rgb_2$ is mounted on the *small beam 1* coordinate frame, which is in turn assembled in the *big beam* coordinate frame. The idea is to define one transformation along the world to sensor chain to be estimated. These selected transformations are highlighted with green arrows in the figure. The other transformations remain unaltered during the calibration. In the example of $rgb_2$, the selected transformation was between the *small beam 1* and *rgb 3* coordinate frames. Note that the selected transformation does not necessarily need to include the sensor's coordinate frame, as is the example of sensor $rgb_2$.

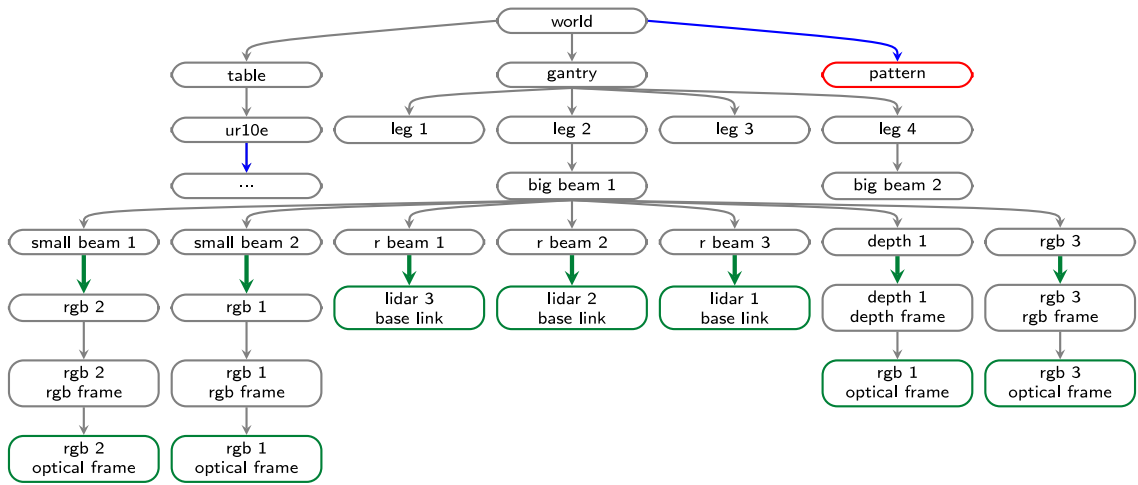

**Fig. 1.** Example of a transformation tree that represents the chain of transformations between coordinate systems of the collaborative cell. Blue arrows signal that transformations are dynamic, green arrows denote that the transformation will be optimized, frames are highlighted in green when sensors output data in that coordinate frame, the red node represents the calibration pattern link, which is both dynamic and is to be calibrated. (For interpretation of the references to color in this figure legend, the reader is referred to the web version of this article.)

The calibration procedure computes the overall sensor to world transformation, i.e. ${}^{s_i}\hat{\mathbf{T}}^w$ in (2), from the chain of transformations for that respective sensor, where one selected transformation is changing during optimization and the others are static.

The process of acquiring data consists of moving a calibration pattern in front of the sensors in a way that the pattern is viewed by all sensors in some moments of the acquisition. The acquisition does not require that the pattern is visible to all sensors at the same time. The dimensions of the calibration pattern must also be specified during the calibration setup.

### 3.2. Automatic and manual labeling

When saving a collection, the sensor data is labeled automatically. That means that information of where the pattern is identified in the data of each sensor is generated automatically. Naturally, the format of these pattern labels differ from modality to modality. For range data, a label is defined as the position of the outer edges of the physical chessboard. For RGB data, a label consists of the pixel coordinates of the inside corners of the pattern.

#### 3.2.1. RGB

The RGB automatic labeling uses the *OpenCV ArUco Marker Detection* toolbox. We have configured that at least 25% of the total number of corners must be identified to assume a valid pattern detection. This automatic labeling of the RGB data using *ChArUcO* patterns is very accurate and efficient. For this reason we have found no need to develop interactive tools to correct the automatic labels and produce manual annotations.

Fig. 2 shows an example of a labeled RGB image.

#### 3.2.2. 3D LiDAR

The label representation for 3D LiDAR data consists of a list of points that belong to the sensor raw data and are identified as viewing the pattern. In addition, we define two separate classes: points that lie on the pattern plane, and a subset of the former that are located on the boundaries of the pattern.

The procedure is semi-automatic, since that it requires the user to define a seed point close to the pattern. This is done using an interactive marker in *Rviz*. That seed is used as the center of a sphere of predefined radius that selects only a small set of points where the pattern should be located. Then, the support plane of the pattern is searched using a *RanSaC* algorithm. Finally, the points that belong to the pattern are obtained as those which are close enough to the support plane, i.e., the *RanSaC* inliers. The boundary points are then found by collecting, for each vertical LiDAR layer, the left and rightmost inliers.

Fig. 3 shows an example of a labeled point cloud, where black points represent the physical limits of the calibration pattern, and green points represent the points inside the calibration pattern.

It must be noted that, if needed, the LiDAR labeling can be reviewed and corrected manually by selecting points in the point cloud and assigning them the proper labels.

#### 3.2.3. Depth camera

The depth camera is labeled using a propagation mechanism, starting from a initial seed point. Similarly to the LiDAR labeling mechanism, the seed point is manually given in the first frame, and from then it is automatically tracked from frame to frame, using the center of mass of the detected pattern in the previous frame. As we are labeling adjacent frames, we can assume that, from frame to frame, the movement of the calibration pattern is small enough so that the center of mass of the previous frame is still inside the area of the pattern in the subsequent frame. The propagation algorithm starts from the initial seed point and uses a tetra-directional flood fill technique to propagate through the area of the calibration pattern. The labels of depth images are separated into two categories: boundary points and inside points. Fig. 4 shows an example of a labeled depth image.

The automatic procedure detailed above does not work accurately in all frames. This is due to the nature of depth images and to the proximity of other objects to the pattern. For this reason, we have developed a dataset reviewer where incorrectly labeled images can be manually annotated by defining a polygon around the pattern. Then, the previously mentioned propagation algorithm is executed with this polygon acting as a propagation constraint, which results in accurately defined labels for depth data.

### 3.3. Calibration

The calibration process reads the previously recorded dataset and loads the labels for each sensor and for each collection into a data structure.

Errors are computed for each sensor modality based on its labels and the position of the chessboard, that are used by the objective function to optimize the transformations. Both the sensors and the position of the pattern have freedom to move their position and rotation. The mindset of also giving freedom of movement to the calibration pattern during calibration outputs better calibration results but has a limitation: the sensors are calibrated in relation to the other sensors and the calibration pattern, but not with the fixed structures of the robotic system. We also propose a way to solve this limitation: optionally, we can anchor one of the sensors, that we know it is in the correct position, and that sensor does not move during calibration, forcing all the other sensors and the calibration pattern to move w.r.t. to the anchored sensor.

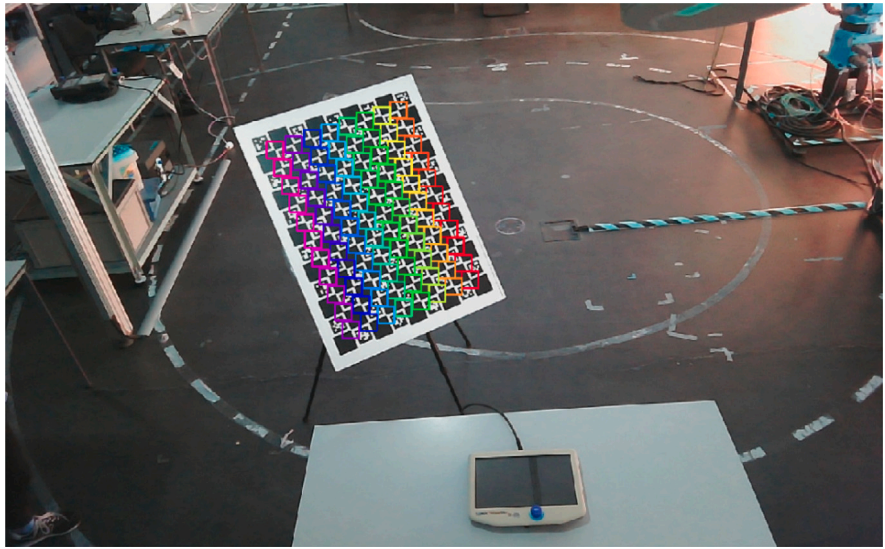

**Fig. 2.** Example of a labeled image in a RGB camera.

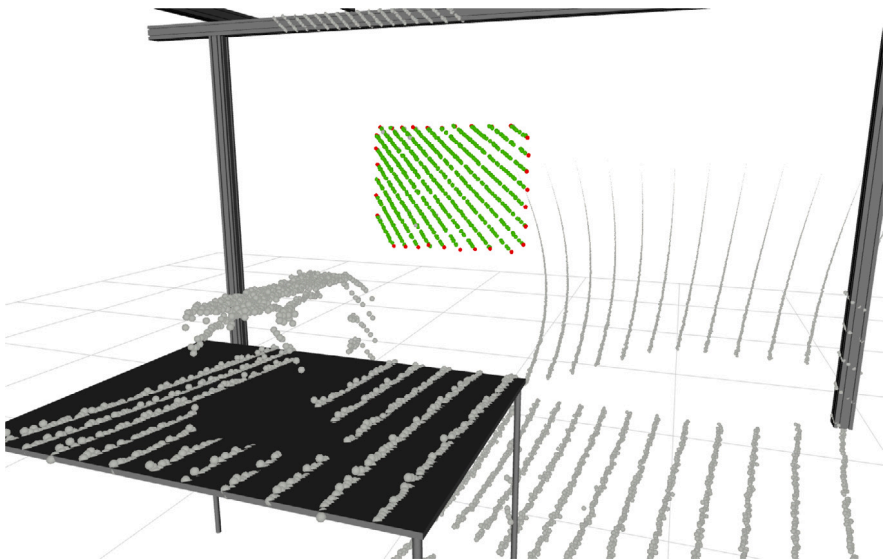

**Fig. 3.** Example of a labeled point cloud from a 3D LiDAR. Gray points are raw data, points annotated as belonging to the pattern are highlighted in green, and points annotated as belonging to the boundaries of the pattern are annotated in red. (For interpretation of the references to color in this figure legend, the reader is referred to the web version of this article.)

### 3.3.1. RGB

Eq. (3) represents the objective function used to compute the error for sensors of the RGB modality. The error per collection, per sensor and per detection, $e_{[c,s,d]}$, is a norm of the difference between the value of the detected label and the projection of the pattern corners in the image:

$$e_{[c,s,d]} = \left\| x_{[c,s,d]} - \mathcal{P}\left( [{}^sT_c^p \times x_d]_{xyz}, k_s, u_s \right) \right\|^2, \tag{3}$$

where $x_{[c,s,d]}$ is the detected label per collection, per sensor and per detection and $\mathcal{P}\left( [{}^sT_c^p \times x_d]_{xyz}, k_s, u_s \right)$ is the projection of the pattern corners, $x_d$, in the image, where $k_s$ and $u_s$ are the intrinsic values of the camera. ${}^sT_c^p$ is the transformation between the pattern and the sensor for that collection.

### 3.3.2. 3D LiDAR

The cost function of 3D LiDARs is divided into two errors: orthogonal error, defined by Eq. (4) and longitudinal error, defined by Eq. (5).

The orthogonal error, $e_{o_{[c,s,d]}}$, is obtained by the $Z$ component of labels projected to the pattern coordinated system:

$$e_{o_{[c,s,d]}} = [({}^sT_c^p)^{-1} \times x_i]_z, \tag{4}$$

where $({}^sT_c^p)^{-1}$ is the inverse transformation between the pattern and the sensor and $x_i$ are the inside points of the detected label.

The longitudinal error, $e_{l_{[c,s,d]}}$, is calculated by the minimum distance between each labeled point projected into the pattern and each

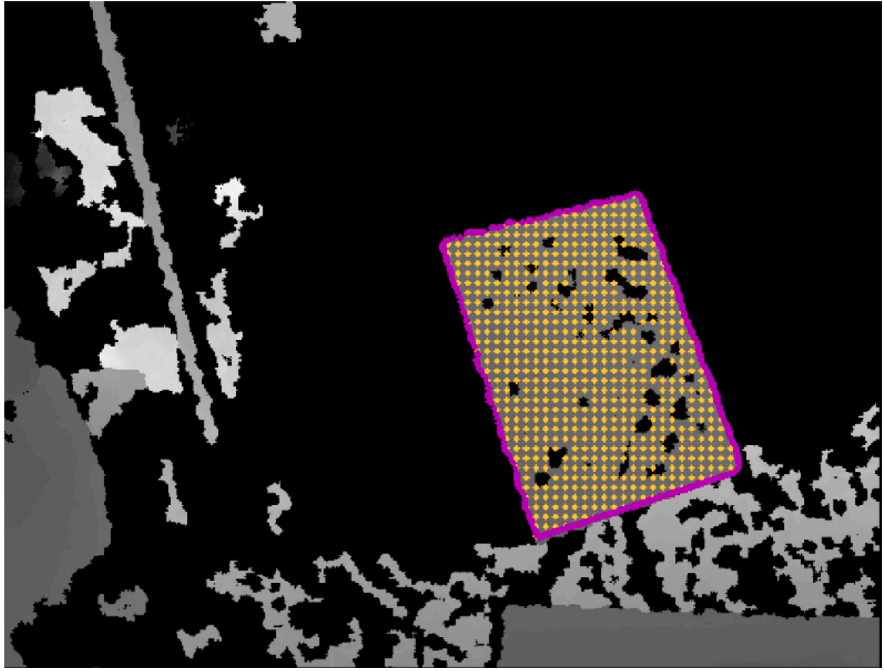

**Fig. 4.** Example of a labeled depth map. Yellow points signal the subsampled lidar points annotated as belonging to the pattern, while purple points denote the annotation of the boundaries of the pattern. (For interpretation of the references to color in this figure legend, the reader is referred to the web version of this article.)

ground truth pattern point:

$$e_{l_{[c,s,d]}} = \min_{q \in Q} \left( \left\| [x_q - (^sT_c^p)^{-1} \times x_b]_{xy} \right\|^2 \right), \quad (5)$$

where $x_q$ is the 3D coordinates obtained by the (spatially periodic) sampling, $(^sT_c^p)^{-1}$ is the inverse transformation between the pattern and the sensor and $x_b$ is the boundary of the detected label.

### 3.3.3. Depth camera

The philosophy under the idealization of the objective function to optimize the depth camera position is essentially the same as the one for LiDARs, since both sensors output range data. The difference is that the labels for depth camera are defined in the image domain and not in the point cloud. As a result of this, and because we know that each pixel value of the depth image is equal to the distance between the camera and that point in the world, it is possible to convert the values in the image to $(X, Y, Z)$ coordinates. Eq. (6) represents that conversion to tridimensional coordinates:

$$\mathcal{F}(x_d) = \begin{cases} Z = image(x_{pix}, y_{pix}) \\ X = \dfrac{x_{pix} - c_x}{f_x} \times Z \\ Y = \dfrac{y_{pix} - c_y}{f_y} \times Z, \end{cases} \quad x_d = (x_{pix}, y_{pix}), \quad (6)$$

where $(f_x, f_y)$ are the bidirectional focal lengths and $(c_x, c_y)$ is the optical center of the image. The point $(X, Y, Z)$ is the 3D coordinate of the label and $x_d$ is the coordinate of the label in pixels.

The orthogonal error, $e_{o_{[c,s,d]}}$ expressed by Eq. (7) is obtained by the $Z$ component of converted labels projected to the pattern coordinate system:

$$e_{o_{[c,s,d]}} = \left[ (^sT_c^p)^{-1} \times \mathcal{F}(x_i) \right]_z, \quad (7)$$

where $(^sT_c^p)^{-1}$ is the inverse transformation between the pattern and the sensor, and $\mathcal{F}(x_i)$ are the inside points of the detected label converted to 3D coordinates.

The longitudinal error, $e_{l_{[c,s,d]}}$, expressed by Eq. (8) is calculated by the minimum distance between each labeled point projected into the pattern and each ground truth pattern point.

$$e_{l_{[c,s,d]}} = \min_{q \in Q} \left( \left\| [x_q - (^sT_c^p)^{-1} \times \mathcal{F}(x_b)]_{xy} \right\|^2 \right), \quad (8)$$

where $x_q$ is the 3D coordinates obtained by the (spatially periodic) sampling, $(^sT_c^p)^{-1}$ is the inverse transformation between the pattern and the sensor and $\mathcal{F}(x_b)$ is conversion of the boundary of the detected label to 3D coordinates.

This error uses only the labeled boundary points. In addition to this, it is also possible to generate a set of points lying on the boundary of the pattern. This is done using the knowledge of the dimensions of the pattern. These are referred to as the ground truth points. The longitudinal error expressed by Eq. (8) is calculated by projecting each labeled point from the sensor coordinate frame to the pattern local coordinate frame (where the ground truth points are defined), and finding the closest ground truth point for each point. Since we are interested only in the longitudinal component of the error, only the XY distance is considered.

## 4. Tests and results

As discussed in previous sections, our approach enables the simultaneous calibration of all the sensors in the system. However, other approaches cannot carry out this global optimization, since they operate with pairs of sensors. Because of this, the assessment of the calibration accuracy is conducted in a pairwise configuration, so that it may be applied both to our methodology (despite the fact that it calibrates the complete system) and also to other approaches.

Tests and results are divided and detailed in the following parts: Collaborative Cell Setup Calibration; RGB to RGB Evaluation; LiDAR to LiDAR Evaluation; LiDAR to RGB Evaluation; LiDAR to Depth Evaluation and Depth to RGB Evaluation.

### 4.1. Collaborative cell setup calibration

As mentioned previously, a collaborative cell is a space where collaborative robots and humans can safely work together. The ultimate goal would be that the robot and human could participate in tasks with a common goal to achieve a more efficient work.

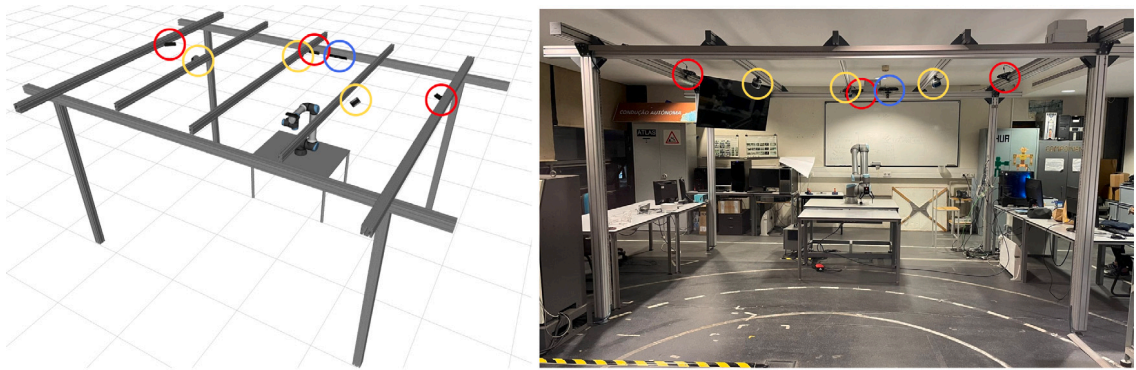

**Fig. 5.** Simulated and real representation of the collaborative cell that serves as case of study. The cell contains a gantry where several RGB, depth and LiDAR sensors are mounted. In the middle of the volume there is table and a robotic manipulator which will interact with human operators. Red circles represent RGB cameras, blue circles represent depth cameras and yellow circles represent 3D LiDAR.

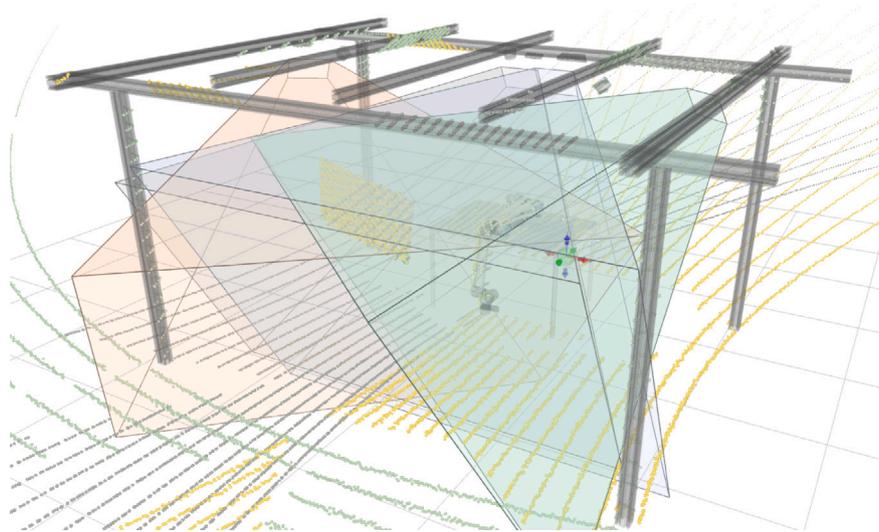

**Fig. 6.** Fields of view of the cameras mounted on the collaborative cell. The point clouds produced by the LiDARs are also shown.

In our particular case, we have built a collaborative cell with 4 m × 2.8 m and 2.29 m high. Fig. 5 shows the collaborative cell in simulated and real environment. In terms of sensors, the cell includes three LiDARs, one RGB-D camera and three RGB cameras. From now on, these sensors will be referred to as $lidar_1$, $lidar_2$ and $lidar_3$, $depth_1$ and $rgb_1$, $rgb_2$ and $rgb_3$.

Fig. 6 shows a representation of the fields of view of the sensors and the coverage of the cell by the LiDAR point clouds. In the image, the gray point clouds come from the $lidar_1$ (right), the green ones from the $lidar_2$ (center) and the yellow ones from the $lidar_3$ (left). The purple frustum represents the field of view of the sensor $depth_1$. The RGB fields of view are represented by the light orange, gray and green for $rgb_1$ (left), $rgb_3$ (center) and $rgb_2$ (right) respectively.

A video[1] has been made available that includes a demonstration of the complete calibration procedure for this collaborative cell.

Table 1 shows the details of the used train and test datasets for the results presented in this section. The train dataset is the dataset that is used for calibration and where the transformations between sensors are estimated. The test dataset is a non-calibrated dataset, with the sensors in the same position as the train dataset, where the results will be evaluated with the transformations obtained in during the calibration of the train dataset.

---

[1] https://youtu.be/KFPUTGR4rBw

**Table 1**
Descriptions of the datasets used in the experiments, where RGB partials mean the number of partial calibration pattern detections in the RGB sensors and complete denotes the number of collections where the calibration pattern was detected by all seven sensors.

| Type of data | Dataset | # collections | # RGB partials | # complete |
|---|---|---|---|---|
| sim. | train dataset | 23 | 35 | 5 |
|  | test dataset | 17 | 26 | 4 |
| real | train dataset | 29 | 61 | 6 |
|  | test dataset | 14 | 29 | 4 |

The evaluation of the calibration is conducted is a pair-wise manner. The results will be presented both in the simulated system and using real data. Note that to calibrate the simulated system, we induced an initial estimate random error of 0.1 m and 0.1 rad to the initial position of the sensors to reinforce the validity of the method in simulation.

### 4.2. RGB to RGB evaluation

The RGB to RGB sensor evaluation is computed by projecting the labels of the source sensor, using the calculated transformation matrix, to the target sensor image and calculating the reprojection errors.

Table 2 shows the root mean square errors for both simulation and real data calibration. In the calibration using simulated data, we obtained sub-pixel accuracy with an average of half a pixel. As expected, the accuracy in real data is lower, with an error of around 1.2 pixels. The reason for this could be that real data is less controlled and has more sources of error than simulation such as, for example, illumination, reflectivity and background noise that might influence the accuracy of detection of the calibration pattern. This table also presents the results using the same data for the *OpenCV Calibration Tool*, a very popular computer vision library used for stereo camera calibration. As discussed in Section 2, most calibration algorithms use a pair-wise methodology, which is the case of *OpenCV Calibration Tool*. This means that, to calibrate the entire system, it would require a sequential pairwise calibration by calibrating all the possible combinations of two sensors. Note that some for camera pairs it was not possible to calibrate using *OpenCV*. This is because these pairs contain camera with a very small overlapping field of view. Moreover *OpenCV* uses a pattern detection that requires that the pattern is fully visible in the image in order to be detected. Because of this, there were no collections in which both cameras in the pair were able to detect the pattern. Since *OpenCV* is a sensor to sensor based approach it cannot operate in these circumstances.

**Table 2**
Pair-wise root mean square errors for the RGB to RGB evaluation in pixels.

| Sensor pair | Our framework | | OpenCV | | Kalibr | |
|---|---|---|---|---|---|---|
| | Sim. | Real | Sim. | Real | Sim. | Real |
| $rgb_1$ to $rgb_2$ | 0.684 | 1.536 | [a] | [a] | [b] | 1.010 |
| $rgb_1$ to $rgb_3$ | 0.463 | 1.085 | 0.675 | 1.828 | 1.290 | 0.906 |
| $rgb_2$ to $rgb_3$ | 0.541 | 1.113 | 0.578 | [a] | [b] | 0.743 |
| average | 0.563 | 1.245 | 0.627 | 1.828 | 1.290 | 0.825 |

[a]OpenCV error: No complete detections of the chessboard.
[b]Kalibr error: Cameras are not connected through mutual observations.

We can also conclude that, even calibrating seven different sensors simultaneously, the proposed approach still managed to obtain better RGB pair-wise results when compared to *OpenCV*.

We have also compared our approach with *Kalibr* [32,42], which is a more recent multi-camera intrinsic and extrinsic calibration tool. We were only able to use *Kalibr* in a pair-wise configuration. This calibration framework, unlike *OpenCV Calibration Tool*, is already a multi-sensor method based on optimization. Nonetheless, this method is not multi-modal and is only able to calibrate RGB cameras. Results are also presented in Table 2.

*Kalibr* also shows the same problem as OpenCV, since that in some situations it is not able to calibrate when cameras have minimal overlapping fields of view. In the real world scenario, *Kalibr* was able to perform calibration in all three camera pairs with sub-pixel performance. Although the performance is slightly better in comparison to our method, it should be noted that this calibration framework is only able to calibrate cameras. In contrast, our system calibrates several modalities all at the same time.

As we can see by looking at Fig. 6, $rgb_1$ (orange frustum) and $rgb_2$ (green frustum) have minimal overlap and very different orientations. This makes it difficult to position the calibration pattern in such a way that it is visible by both sensors. For that reason, that camera pair is the most difficult to detect. Unlike OpenCV, *Kalibr* can calibrate this pair in the real data scenario because it uses a different chessboard detector which does not require that the pattern is fully visible in the image. Even so, of the 29 available collections in the training real dataset, *Kalibr* was only able to use 8 for calibration.

Regarding the $rgb_2 - rgb_3$ pair, *OpenCV* was not able to calibrate with the real data because the detections of camera $rgb_2$ are all partial, which is a problem that *Kalibr* does not have. Regarding simulation, camera $rgb_2$ has more complete detections using our pattern detection algorithm. However, the *Kalibr* detection algorithm was not able to produce detections of the pattern for both images in the same collection, and for that reason, it was not able to calibrate.

### 4.3. LiDAR to LiDAR evaluation

The evaluation between LiDAR pairs is conducted by transforming the points of the source LiDAR into the coordinate system of the target LiDAR. Then, for each point in the target LiDAR, we obtain the closest transformed point in the source LiDAR and compute the distance between both.

Table 3 shows the calibration errors for the LiDAR-LiDAR pairs. Considering that the calibration pattern is at a distance of 2–2.5 meters from each LiDAR, the maximum distance between LiDAR points measuring the pattern is around 100 mm. When transforming the labeled points of the source LiDAR to the target LiDAR coordinate system, the labels could end up in such a way that the scan of the two LiDARs have a significant displacement between them caused by the low sensor resolution.

Considering this low sensor resolution it is natural that the error values in Table 3 are in the magnitude of a few tenths of millimeters. Table 3 also shows the calibration results for the same datasets using the Iterative Closest Point (ICP) algorithm. ICP is a very common iterative solution for the alignment of two sets of 3D data. The difference between **Initial** and **Aligned** ICPs indications in the table is that **Initial** has the same initial estimate as our framework, and **Aligned** has a better initial estimate created by manually aligning the point clouds. The ICP algorithm is executed for the pair of points clouds in each collection, which means that, as *OpenCV Calibration Tool*, it is also a pair-wise algorithm and requires some form of sequential pairwise calibration to calibrate all of the sensors in the system. Thus, there is an estimated transformation for each collection. The difference between the **Best** and **Average** ICPs is that the **Average** uses the average transformation estimated for all collections, while the **Best** makes use of the estimated transformation which had the least amount of estimated ICP error.

When comparing the ICP results with our framework, we can conclude that the only one that comes close is the ICP Aligned Average. Nevertheless, the ICP is a pair-wise method, while our method obtained similar results while calibrating all sensors simultaneously.

### 4.4. LiDAR to RGB evaluation

The LiDAR to RGB camera error metric is assessed by projecting the LiDAR labeled points to the RGB image using the transformation between sensors estimated during calibration.

However, the labels of the RGB data correspond to the inside corners of the chessboard or *ChArUcO*. In contrast, the LiDAR data labels correspond to the physical limits of the chessboard. Therefore, the RGB images for each collection need to be manually labeled in the test dataset to identify the physical limits of the pattern. Those labels can then be compared to the LiDAR labels using the reprojection error.

Table 4 shows the reprojection errors obtained from pair-wise evaluations of the calibration of both the simulated and the real data. As explained before, the LiDARs have a low resolution so it is expected that these errors have higher magnitude when compared with RGB to RGB evaluation errors. In this evaluation we can see that results are on average around 1 or 2 pixels of reprojection error. The difference between simulated and real results is approximately 0.5 pixels. This shows that the evaluation is consistent: as expected the real data is less accurate.

Also, there is no significant difference between the several pairs of sensors. Our explanation is that since the LiDARs have all around fields of view there is complete overlap between all camera-LiDAR pairs.

Fig. 7 shows the projection of the three LiDARs point clouds into an RGB frame after calibration. The point clouds are colored according to the distance to each sensor. As such, changes in object in the image should align which changes in color of the points clouds. As we can see, point clouds align almost perfectly with the shape of the chessboard in the image, which proves that the calibration was successful.

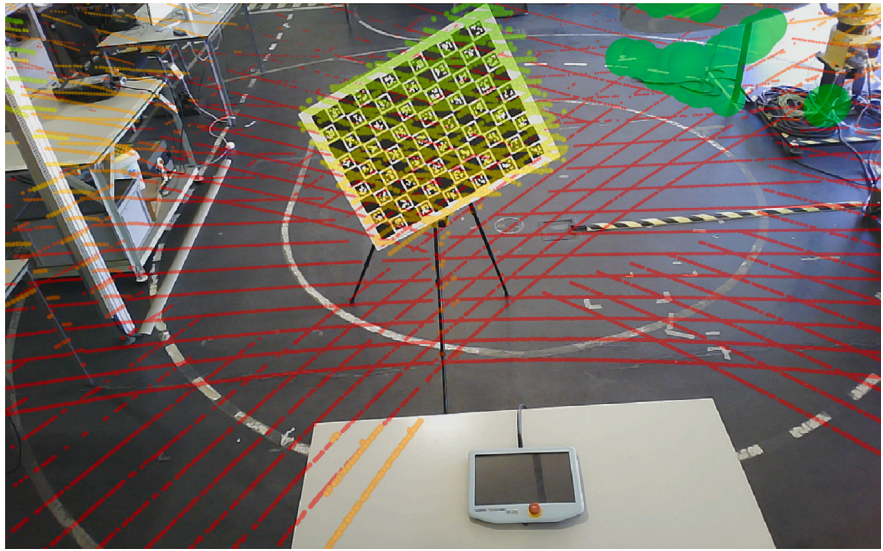

**Fig. 7.** Projection of point clouds from all LiDARs to the image of camera $rgb_3$ after calibration. The point clouds are colored according to the distance to each sensor. As such, changes in object in the image should align which changes in color of the points clouds.

**Table 3**
Pair-wise root mean square errors for the LiDAR to LiDAR sensors evaluation in millimeters.

| Sensor pair | Our framework | | ICP initial average | | ICP initial best | | ICP aligned average | | ICP aligned best | |
|---|---|---|---|---|---|---|---|---|---|---|
| | Sim. | Real | Sim. | Real | Sim. | Real | Sim. | Real | Sim. | Real |
| $lidar_1$ to $lidar_2$ | 26.472 | 77.504 | 76.541 | 294.977 | 28.088 | 207.025 | 30.398 | 68.524 | 127.468 | 75.617 |
| $lidar_1$ to $lidar_3$ | 33.049 | 69.234 | 248.526 | 84.633 | 43.230 | 265.582 | 33.033 | 67.312 | 32.367 | 70.103 |
| $lidar_2$ to $lidar_3$ | 39.401 | 13.946 | 423.132 | 140.611 | 38.212 | 15.900 | 38.361 | 24.105 | 115.198 | 183.892 |
| average | 32.974 | 53.561 | 249.400 | 173.407 | 36.510 | 162.836 | 33.931 | 53.314 | 91.678 | 109.871 |

**Table 4**
Pair-wise root mean square errors for the LiDAR to RGB sensors evaluation in pixels.

| Sensor Pair | Sim. | Real |
|---|---|---|
| $lidar_1$ to $rgb_1$ | 1.726 | 2.516 |
| $lidar_2$ to $rgb_1$ | 1.801 | 2.582 |
| $lidar_3$ to $rgb_1$ | 2.692 | 3.297 |
| $lidar_1$ to $rgb_2$ | 3.502 | 3.837 |
| $lidar_2$ to $rgb_2$ | 3.659 | 3.173 |
| $lidar_3$ to $rgb_2$ | 2.854 | 2.754 |
| $lidar_1$ to $rgb_3$ | 2.892 | 3.767 |
| $lidar_2$ to $rgb_3$ | 1.763 | 2.768 |
| $lidar_3$ to $rgb_3$ | 2.352 | 3.189 |
| Average | 2.582 | 3.098 |

*4.5. LiDAR to depth evaluation*

Similarly to the LiDAR-RGB evaluation, the LiDAR-depth evaluation consists of projecting the LiDAR points to the depth image. The difference is that the depth labels are also the physical limits of the chessboard, so we can directly compare the points without needing additional manual labeling.

Table 5 shows the results of the calibration error for simulated and real data. Once again, simulated and real results have a small sub-pixel difference which shows consistency.

Table 5 also shows calibration results using the ICP technique, where the different variants are the same as the ones in the LiDAR-LiDAR evaluation. None of these techniques obtains calibration results as good as the ones obtained with our methodology.

Fig. 8 shows the projection of the three LiDAR point clouds into the depth map. As we can see, the calibrated point clouds align well with the pattern and other features in the image, like the table at the bottom, and the structure of the cell in the left side.

*4.6. Depth to RGB evaluation*

On the depth to RGB pair-wise evaluation, we project the depth labels to the RGB image using the transformations obtained during calibration. Again, there is a difference between the nature of the labels, so we use the annotations of the RGB images that were already made for the LiDAR-RGB evaluation to compare the physical pattern limits.

Table 6 shows the calibration errors of the depth-RGB pairs. The average errors are around 3 pixels, which are clearly above those for the LiDAR to RGB evaluation. We believe this is because the depth estimation is very as precise in the depth sensors when compared to LiDARs.

## 5. Conclusions and future work

This paper solves the problem of the calibration of complex, multi-sensor and multi-modal systems. To do so, we created a calibration framework based on a sensor to pattern paradigm, which has clear advantages over sensor to sensor calibrations, which are the basis for most of the current calibration approaches. Our approach provides several improvements w.r.t. the state-of-the-art, such as:

- a solution to calibrate any number of sensors and several modalities;
- a solution for systems with non-overlapping fields of view;
- the ability to accurately calibrate RGB cameras with partial detections;
- the simultaneous calibration of any number of sensors;

Furthermore, we provide a complete calibration framework with seamless integration with the Robot Operating System (ROS) ecosystem, available at https://github.com/lardemua/atom.

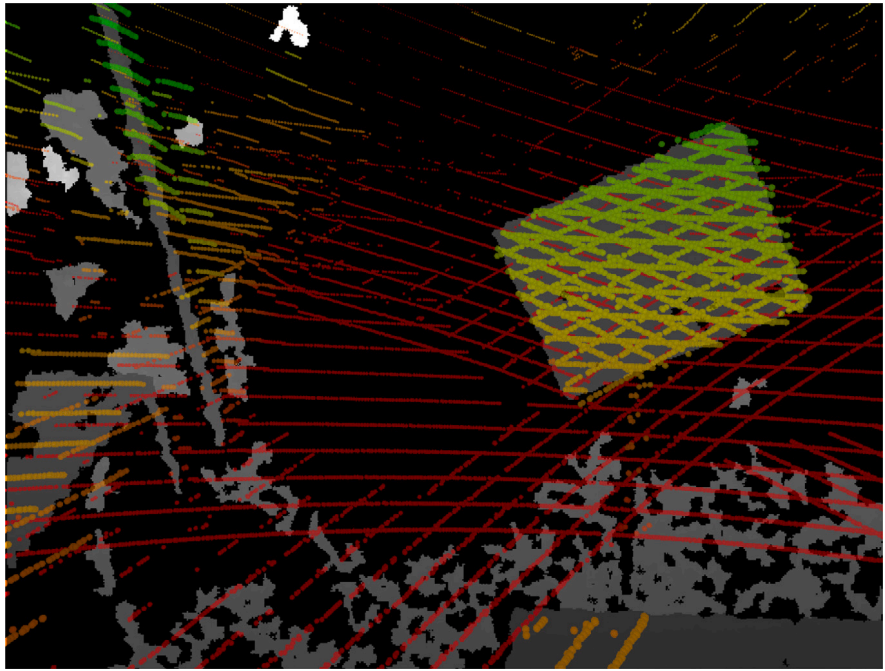

**Fig. 8.** Projection of point clouds in the $depth_1$ sensor depth map after calibration. The point clouds are colored according to the distance to each sensor. As such, changes in object in the image should align which changes in color of the points clouds.

**Table 5**
Pair-wise root mean square errors for the LiDAR to depth sensors evaluation in pixels.

| Sensor pair | Our framework | | ICP initial average | | ICP initial best | | ICP aligned average | | ICP aligned best | |
|---|---|---|---|---|---|---|---|---|---|---|
| | Sim. | Real | Sim. | Real | Sim. | Real | Sim. | Real | Sim. | Real |
| $lidar_1$ to $depth_1$ | 1.281 | 1.791 | 12.591 | 166.459 | 1.575 | 84.895 | 5.563 | 9.170 | 2.094 | 8.339 |
| $lidar_2$ to $depth_1$ | 1.054 | 1.608 | 16.478 | 30.654 | 45.970 | 4.883 | 2.477 | 4.392 | 1.915 | 6.114 |
| $lidar_3$ to $depth_1$ | 1.584 | 2.058 | 34.246 | 148.168 | 5.951 | 144.052 | 3.025 | 2.055 | 3.493 | 2.050 |
| Average | 1.306 | 1.819 | 21.105 | 115.094 | 17.832 | 77.943 | 2.751 | 5.206 | 2.501 | 5.501 |

**Table 6**
Pair-wise root mean square errors for depth to RGB sensors evaluation in pixels.

| Sensor Pair | Sim. | Real |
|---|---|---|
| $depth_1$ to $rgb_1$ | 3.328 | 3.990 |
| $depth_1$ to $rgb_2$ | 3.212 | 4.553 |
| $depth_1$ to $rgb_3$ | 3.642 | 3.584 |
| Average | 3.394 | 4.042 |

Results show that our framework is able to achieve similar, or even better performance when compared with other state-of-the-art pair-wise calibration methods, while calibrating all sensors from three different modalities simultaneously.

One shortcoming of our approach is the inability to calibrate the sensors with the structure of the robotic system. For example, in the case of the collaborative cell used in the experiments, it was necessary to manually calibrate one sensor w.r.t. the gantry structure. Then, this sensors is fixed and the calibration moves all other sensors w.r.t. the fixed one. A better, automatic procedure for solving this problem would be an interesting addition.

As discussed throughout the paper, collaborative cells are highly complex systems which render current calibration approaches unusable. Furthermore, collaborative cells have several additional challenges, such as the minimal overlapping fields of view between sensors. Our approach is able to tackle all these challenges, as the method was able to carry out a successful calibration of a highly complex collaborative cell.

Future work includes extending our framework to more modalities, like thermal cameras. We also intend to extend the calibration of our collaborative cell by installing an RGB-D camera in the end-effector of the robotic arm, as well as including a hand-eye configuration in the set of calibration challenges.

**CRediT authorship contribution statement**

**Daniela Rato:** Conceptualization of this study, Methodology, Software, Writing – original draft. **Miguel Oliveira:** Conceptualization of this study, Methodology, Software, Writing. **Vítor Santos:** Conceptualization of this study, Methodology, Writing – review & editing, Funding. **Manuel Gomes:** Software, Validation and results. **Angel Sappa:** Writing – review & editing.

**Declaration of competing interest**

The authors declare that they have no known competing financial interests or personal relationships that could have appeared to influence the work reported in this paper.